%% file: main.tex
\documentclass[10pt,twocolumn,letterpaper]{article}

\usepackage{cvpr}              % To produce the CAMERA-READY version
\input{preamble}

\definecolor{cvprblue}{rgb}{0.21,0.49,0.74}
\usepackage[pagebackref,breaklinks,colorlinks,allcolors=cvprblue]{hyperref}
\usepackage{multirow}
\usepackage{colortbl}
\usepackage{diagbox}
\usepackage[accsupp]{axessibility} 
\def\paperID{18406} % *** Enter the Paper ID here
\def\confName{CVPR}
\def\confYear{2026}

\title{SDDF: Specificity-Driven Dynamic Focusing for Open-Vocabulary Camouflaged Object Detection
}

\author{
    Jiaming Liang$^{1,2,*}$, 
    Yifeng Zhan$^{1,*}$, 
    Chunlin Liu$^{1}$, 
    Weihua Zheng$^{1}$, 
    Bingye Peng$^1$ \\
    Qiwei Liang$^{1,2}$, 
    Boyang Cai$^1$, 
    Xiaochun Mai$^{1,\dagger}$, 
    Qiang Nie$^{2,\dagger}$ 
    \vspace{8pt} \\ % 增加姓名与学校之间的间距
    $^1$Shenzhen University \quad
    $^2$The Hong Kong University of Science and Technology (Guangzhou) \\
    {\small $^*$Equal contribution \quad $^\dagger$Corresponding author} \\
}

\begin{document}
\maketitle
\begin{abstract}
%开放词汇目标检测旨在通过文本提示去检测开放世界中已知或未知的目标。得益于大规模预训练视觉语言模型的出现，OVOD展现出不俗的零样本泛化能力。然而，面对具有伪装性的物体，由于目标与背景视觉特征高度相似，导致检测器难以区分定位目标。为了弥补这一研究空白，我们在筛选的伪装目标图像数据上融入细粒度文本描述，构建了命名为OVCOD-D的benckmark。受限于伪装目标图像数据的规模，我们选择在大规模目标检测数据集预训练的检测器作为基准方法，其具有更强的零样本泛化能力。在多模态大模型生成的特异性子描述中，仍存在易混淆的修饰性信息，因此我们设计子描述主成分对比融合策略，来减少干扰性文本描述。面对伪装目标视觉特征与周围环境高度相似的困境，我们提出特异性引导的区域弱对齐与动态聚焦方法，旨在强化检测器区分伪装目标和背景物体的能力。在zero-shot测试上，我们提出的方法在OVCOD-D上达到49.3的AP。
Open-vocabulary object detection (OVOD) aims to detect known and unknown objects in the open world by leveraging text prompts.
Benefiting from the emergence of large-scale vision--language pre-trained models, OVOD has demonstrated strong zero-shot generalization capabilities. However, when dealing with camouflaged objects, the detector often fails to distinguish and localize objects because the visual features of the objects and the background are highly similar. To bridge this gap, we construct a benchmark named \textbf{OVCOD-D} by augmenting carefully selected camouflaged object images with fine-grained textual descriptions. Due to the limited scale of available camouflaged object datasets, we adopt detectors pre-trained on large-scale object detection datasets as our baseline methods, as they possess stronger zero-shot generalization ability. In the specificity-aware sub-descriptions generated by multimodal large models, there still exist confusing and overly decorative modifiers. To mitigate such interference, we design a sub-description principal component contrastive fusion strategy that reduces noisy textual components. Furthermore, to address the challenge that the visual features of camouflaged objects are highly similar to those of their surrounding environment, we propose a specificity-guided regional weak alignment and dynamic focusing method, which aims to strengthen the detector’s ability to discriminate camouflaged objects from background. Under the open-set evaluation setting, the proposed method achieves an AP of 56.4 on the OVCOD-D benchmark. Project code and benchmark will be released at \url{https://github.com/Zh1fen/SDDF}.

\end{abstract}

% \begin{figure}[t]
%     \centering
%     \includegraphics[width=\columnwidth]{figure/figure1.pdf}
%     \caption{} 
%     % \yong{Yong: use another example that is different from the one in Fig 3?}}
%     \label{Fig.1}
% \vspace{-9 pt}
% \end{figure}

\section{Introduction}
\label{sec:intro}

%目标检测，作为计算机视觉的一项基本任务，在多年研究中已取得长足的进展，特别是CNN架构的发明大幅提升目标检测模型的性能。相比闭集目标检测器只能在已知类别目标域中进行多目标检测，开放词汇目标检测器可通过目标类别文本提示来检测未知类别的目标，更加灵活应对复杂多变的环境。得益于大规模图像文本对预训练的视觉语言模型（CLIP）的成功，跨模态对齐通过文本语义提示关联到物体视觉特征，让模型具备更强的零样本泛化能力。感谢最前沿工作提出的，基于YOLO架构的轻量化开放词汇目标检测器（YOLO_World、DOSOD、YOLOE），在维持较高检测性能和零样本能力的同时，大幅提高了模型的推理速度，使其能在边端设备部署，推动开放词汇目标检测模型在场景理解、自动驾驶、机器人感知等领域的应用落地。然而，开放词汇目标检测在现实世界中的一大挑战是未知类别目标特征与背景及其相似，难以进行伪装目标检测。针对该挑战，本文介绍一个新任务- -开放词汇伪装目标检测，并在伪装目标图像数据集中融入细粒度文本描述，构建了一个开放词汇伪装目标检测benchmark。

%【第二段写已有伪装目标检测的SOTA方法及其优缺点。】

%直觉来说，直接利用有限的伪装目标数据训练模型，不利于模型学习空间物体的基本表征，会抑制模型的零样本泛化潜质。因此，我们选择在大规模目标检测数据集上预训练的检测器作为基准方法。其次，现有的伪装目标检测模型大都选择参数量大的backbone，检测头也主要用于实例分割，与现实应用需求不吻合。因此，我们基于YOLO架构去设计模型，目的是缩短研究到应用的gap。在视觉文本跨模态学习过程中，多模态大模型生成的细粒度目标描述仍存在混淆的修饰性信息，会错误引导视觉判别性学习。于是我们设计子描述主成分对比融合策略，通过奇异值分解去除干扰性描述，利用子描述与目标和背景的差异，进行对比融合来保留目标描述的特异多元性。面对伪装目标视觉特征和背景高度相似的困境，我们希望通过细粒度语义先验来促进目标和背景间的区分。首先我们提出特异性区域弱对齐策略，利用覆盖率损失使特异性区域向GT区域靠近，增强特异性子描述和目标视觉区域的联系。其次，我们提出空间聚焦GLU，利用特异性目标描述动态增强目标域视觉特征响应，强化模型区分伪装目标和周围环境的能力。实验结果表明，在零样本测试中，我们的方法在OVCOD-D上高达49.3 AP，同时在闭集伪装目标检测上也表现出优异性能。我们的贡献可以总下为如下几点：
%1、构建全新的开放词汇伪装目标检测benckmark，整合优化主流伪装目标图像数据，并注入经过筛选的细粒度伪装目标描述；
%2、我们设计子描述主成分对比融合策略，以及提出特异性引导的空间弱对齐和动态聚焦方法，有效利用细粒度语义先验增强模型定位和区分伪装目标的能力；
%3、大量的实验结果表明，我们的模型在开集和闭集的开放词汇伪装目标检测展现出remarkable性能，也足够轻量化使其可以在端侧部署。

\begin{figure}
    \centering
    \includegraphics[width=\linewidth]{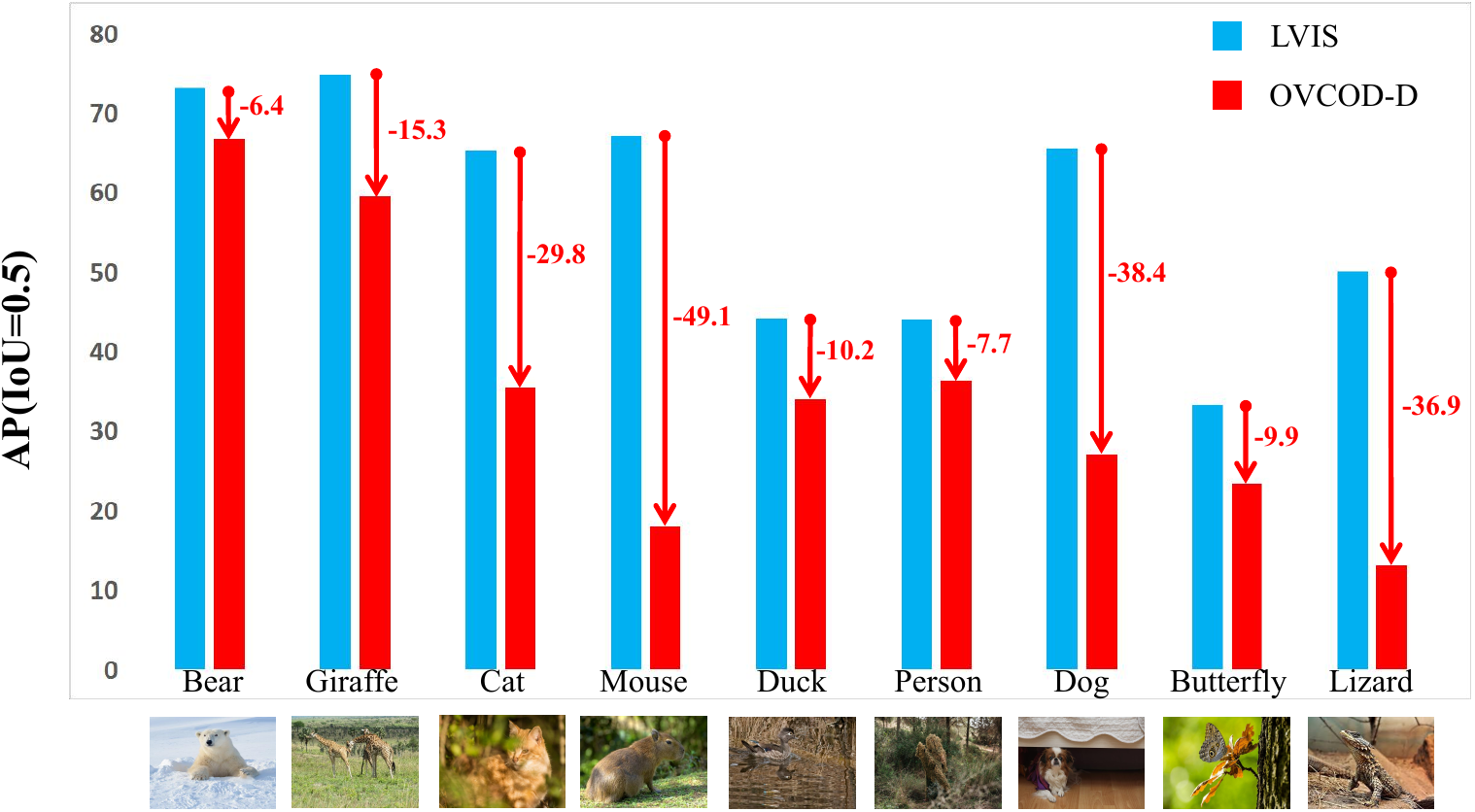}
    \caption{We perform zero-shot detection with the YOLO-World-M~\cite{cheng2024yolo} on both the LVIS~\cite{gupta2019lvis} dataset and our OVCOD-D dataset. By comparing the AP of the overlapping categories across the two datasets, we observe a substantial performance decline on OVCOD-D, indicating that open-vocabulary detectors face significant challenges when dealing with camouflaged objects.}
    % 我们使用YOLO-World模型分别对LVIS数据集和OVCOD-D数据集进行零样本检测，并统计两个数据集中相同类别的AP进行比较，发现开放词汇检测器在OVCOD-D上的性能下降显著
    \label{fig:ovod vs ovcod}
\end{figure}

Object detection, as a fundamental task in applications of scene understanding, autonomous driving and robotic perception, has made remarkable progress over the years~\cite{he2017mask, lin2017feature, redmon2016you, ren2015faster, sandler2018mobilenetv2, zhang2025review}. Unlike closed-set object detectors~\cite{zou2023object,10533619,girshick2014rich,girshick2015fast} that can only identify multiple objects within a predefined category space~\cite{wang2021salientobjectdetectiondeep}, open-vocabulary object detectors are capable of recognizing unseen object categories guided by textual prompts. This enables them to adapt more flexibly to complex and dynamic environments. Benefiting from the success of vision-language models~\cite{radford2021learning} pre-trained on large-scale image-text pairs, cross-modal alignment enables the association of textual semantic cues with visual object features, thereby endowing the model with stronger zero-shot generalization capability~\cite{li2022grounded, liu2024grounding, yao2022detclip}.

\begin{figure}
    \centering
    \includegraphics[width=\linewidth]{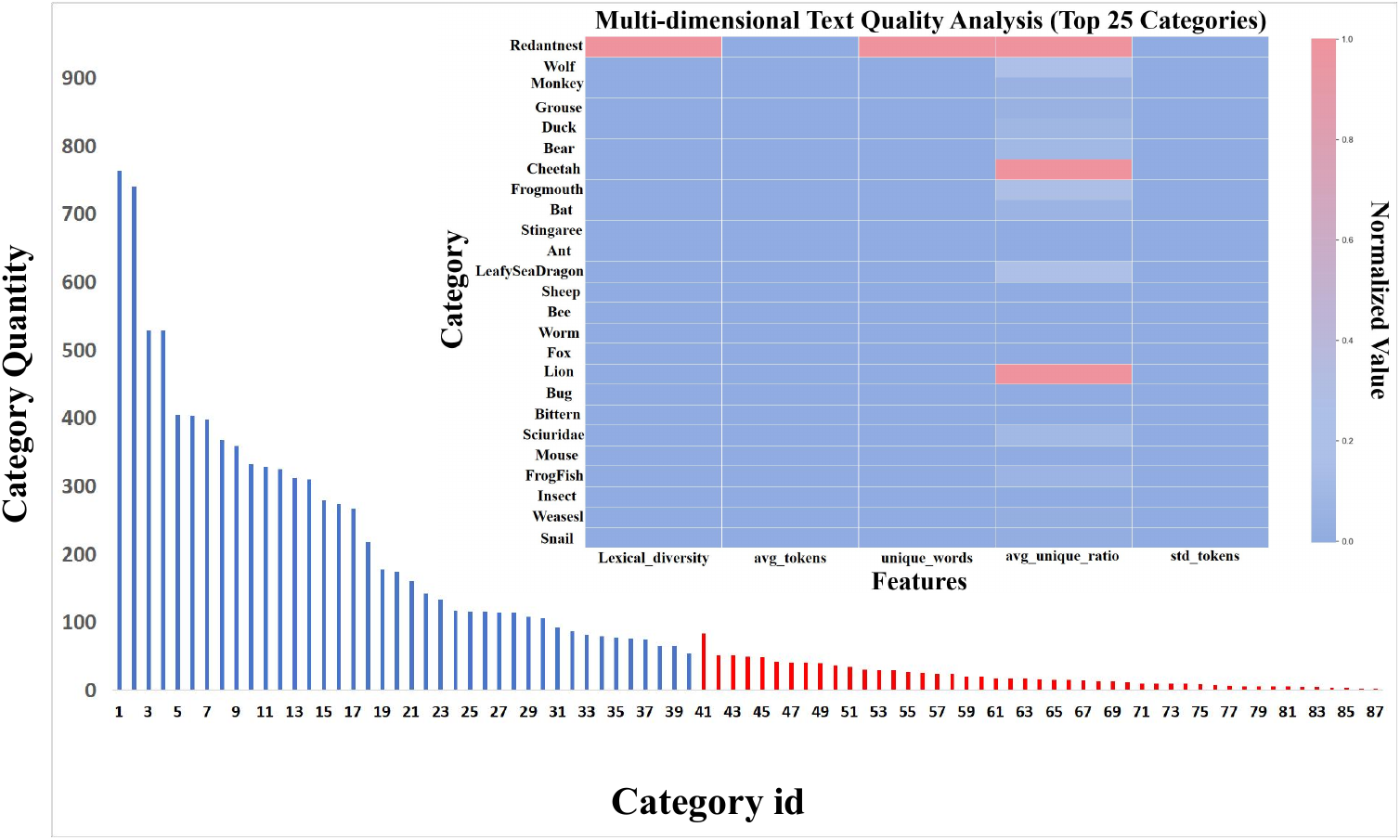}
    \caption{We conducted a statistical analysis of per-class instance counts in the OVCOD-D dataset. In the bar chart, \textcolor{blue}{blue bars} denote base classes and \textcolor{red}{red bars} denote novel classes, revealing a pronounced long-tailed distribution of class frequencies. Additionally, using the fine-grained textual descriptions associated with each category, we selected 25 categories characterized by higher lexical richness to construct a multidimensional quality-analysis heatmap. The horizontal axis reports, in sequence, lexical diversity, average tokens, unique words, average unique-word ratio, and the standard deviation of sentence length.
}

% 我们对OVCOD-D数据集中每个类别数量进行了统计，柱状图中的蓝色部分为基础类，红色部分为新类，可以见到数据集中的类别数量呈现长尾分布的趋势。同时我们也根据每个种类的细粒度文本描述，选择了词汇丰富度较高的15个类别生成了多维度质量分析热力图，横轴依次展示了词汇多样性、平均词数、唯一词数、平均唯一词比例和句子长度标准差。
    \label{fig:Category_quantity_token}
\end{figure}

Even though recent pioneering works in OVOD, built upon the light-weight YOLO architecture~\cite{cheng2024yolo, liu2024yolo, he2025light, wang2025yoloe}, have achieved an impressive balance between detection accuracy, zero-shot capability, and inference efficiency. As illustrated in Figure~\ref{fig:ovod vs ovcod}, OVOD exhibits limited capability in localizing common camouflaged objects, leading to substantial performance degradation under camouflage conditions. Existing OVOD methods struggle with two issues on camouflaged objects: 1) \textit{Redundancy in textual embeddings}, indicated by the low lexical diversity and avg\_unique\_ratio in Figure~\ref{fig:Category_quantity_token}, the fine-grained descriptions from large multimodal models contain superfluous modifiers. In cross-modal vision-language learning, these redundant semantics can introduce noise, thereby misguiding the extraction of discriminative visual features. 2) \textit{Highly similar object and background embeddings issue}, that is, the visual characteristics of camouflaged objects are highly similar to those of the background, resulting in difficult decision boundary learning for camouflaged objects and background categories in the embedding space.
%开放词汇伪装目标检测的难点

% For this task, we construct a dedicated benchmark by integrating fine-grained textual descriptions into existing camouflaged object datasets.

% Intuitively, directly training a model solely on limited camouflaged-object data is not conducive to learning fundamental spatial object representations and can severely hinder its zero-shot generalization potential. \textcolor{blue}{Therefore,} we adopt detectors pre-trained on large-scale object detection datasets as our baseline, leveraging their zero-shot transferability. Second, most existing camouflaged object detection models employ heavy backbones with a large number of parameters, and their detection heads are primarily tailored for instance segmentation, which does not align well with practical application requirements. In contrast, we design our model based on the YOLO architecture with the explicit goal of narrowing the gap between research and real-world deployment. %这段放在第3部分里

To address the above issues, we introduce a new task setting, i.e., Open-Vocabulary Camouflaged Object Detection (OVCOD), constructing a textual description-included \emph{OVCOD-D} benchmark and proposing a method named \emph{SDDF}. To alleviate the limitation of redundant textual-description embeddings, we design a {\emph{sub-description principal component contrastive fusion}} strategy, which first removes interfering textual components via singular value decomposition (SVD), and then exploits the contrastiveness between the sub-description principal components with respect to the object and background regions to perform fusion, thereby preserving the specific and diverse components of sub-descriptions for camouflaged objects.
Furthermore, to tackle the issue of highly similar object and background embeddings, we exploit fine-grained semantic priors to facilitate the discriminant learning between objects and background. Specifically, we introduce a \emph{specific region weak alignment} method, which employs a coverage-based loss to encourage the regions with object-specificity to move closer to the ground-truth (GT) object regions, thereby strengthening the correspondence between specificity-aware textural descriptions and visual regions of the object. In particular, we design a \emph{Spatially Focused-Gated Linear Unit (SF-GLU)} layer, which dynamically enhances the visual feature responses in the object domain under the guidance of object sub-descriptions, thus reinforcing the model’s ability to distinguish camouflaged objects from their surroundings.

Experimental results demonstrate that, under the open-set setting, our method achieves an AP of 56.4 on the OVCOD-D benchmark, while also exhibiting strong performance on closed-set camouflaged object detection. Our main contributions can be summarized as follows:
\begin{itemize}
    \item We construct an open-vocabulary camouflaged object detection benchmark, OVCOD-D, by integrating and refining mainstream camouflaged object image datasets and injecting carefully curated fine-grained descriptions of camouflaged objects.
    \item We devise a sub-description principal component contrastive fusion strategy, while proposing a specificity-guided regional weak alignment and dynamic focusing scheme, which together exploit fine-grained semantic priors to effectively enhance the model’s ability to localize and distinguish camouflaged objects.

    \item Extensive experimental results demonstrate that our model exhibits remarkable performance in both open-set and closed-set camouflaged object detection, while remaining sufficiently lightweight for deployment on edge devices.

\end{itemize}

\section{Related Work}
\label{sec:related_work}

\subsection{Open-Vocabulary Object Detection}

Open-vocabulary object detection (OVOD) \cite{zhu2024surveyopenvocabularydetectionsegmentation} aims to move beyond the reliance of traditional closed-set detectors on predefined category lists, enabling the recognition and localization of previously unseen classes. Early approaches\cite{wang2018zero,hsieh2019one} primarily followed the zero-shot learning paradigm, performing cross-modal matching between pretrained visual features and semantic embeddings\cite{mikolov2013efficient,pennington2014glove}. However, the coarse granularity of these semantic representations limits their capacity to distinguish fine-grained categories.

In recent years, the development of large-scale vision–language pretrained models~\cite{radford2021learning,li2021align,jia2021scaling,singh2022flava,jin2024llms,chen2022open} has ushered in a new phase for OVOD. Detic~\cite{zhou2022detecting} introduces a class-agnostic detection head and prompt integration strategy, substantially narrowing the gap between rare and frequent categories. 
% GLIP~\cite{li2022grounded,xie2025fgclipfinegrainedvisualtextual,faghri2025mobileclip2improvingmultimodalreinforced,Cherti_2023,xu2024demystifyingclipdata}
GLIP~\cite{li2022grounded} presents grounding paradigm effectively addresses challenges posed by similar categories and partially occluded objects, surpassing contemporary baselines. YOLO-UniOW~\cite{liu2024yolo} achieved lightweight cross-modal alignment via adaptive decision learning (AdaDL), it offers an efficient solution for deployment in dynamic scenarios. However, these open-vocabulary detection methods still exhibit notable limitations: most studies pay insufficient attention to complex visual phenomena such as camouflage and occlusion. In addition, the field lacks open-vocabulary datasets specifically tailored to camouflaged object detection (COD), hindering rigorous evaluation of models’ open-vocabulary capabilities in fine-grained camouflage scenarios. 

To tackle this challenge, we construct a fine-grained open-vocabulary object detection dataset for COD. The dataset encompasses a rich spectrum of camouflaged categories and scenes and provides fine-grained descriptions, thereby enhancing the learning and detection of camouflaged objects by vision–language models in open-world contexts and furnishing essential data support for the practical deployment of dynamic detection of multiple camouflaged objects in real-world scenarios.

%Related Work 第一部分  开放词汇目标检测（Open-Vocabulary Object Detection）
% 开放词汇目标检测旨在突破传统闭集检测对预定义类别集合的依赖，实现对未见过类别目标的识别与定位。早期方法主要基于零样本学习（Zero-Shot Learning）框架，通过预训练的视觉特征与语义嵌入（如Word2Vec、GloVe）进行跨模态匹配，但受限于语义表征的粗粒度，对细分类别的区分能力较弱。
% 近年来，随着大规模视觉-语言预训练模型的发展，开放词汇检测进入新阶段。Detic[Zhou et al., 2022]创新性提出类别无关检测头与提示集成策略，显著缩小了稀有类别与常见类别的性能差距。GLIP[Li et al., 2022]提出的phrase grounding范式有效解决了相似类别和部分遮挡目标的检测难题。YOLO-UniOW[Li et al., 2025]首次统一开放词汇与开放世界检测任务，通过自适应决策学习（AdaDL）实现轻量级跨模态对齐，为动态场景落地提供了高效解决方案。然而，现有开放词汇检测方法仍存在局限：多数对伪装、遮挡等复杂视觉现象的关注不足，同时缺乏面向伪装目标检测（COD）领域的专用开放词汇数据集，难以有效评估模型在细粒度伪装场景下的开放词汇检测能力。
% 针对这一挑战，我们构造出面向COD领域的细粒度开放词汇目标检测数据集，该数据集不仅涵盖丰富的伪装目标类别与场景，还提供细粒度的目标描述信息，有助于增强开放世界中视觉语言模型对伪装目标的学习和检测，为现实中动态多伪装目标检测的落地实现提供关键数据支撑。

%第四版 修改者  郑纬华   时间  2025/11/10

\subsection{Camouflaged Object Detection}
Recent years have witnessed the proposal and development of numerous deep learning-based Camouflaged Object Detection (COD) models \cite{LIANG2024127050,Camouflage20CVPR,21PAMI-Concealed}, with recent research increasingly focusing on uncertainty estimation in this field. 
% These deep learning-driven COD approaches enable the automatic learning of intricate discriminative features through neural networks and can be categorized based on three core dimensions: network architectures (i.e., linear, aggregative, branched, and hybrid types, with backbone selections primarily limited to two categories—convolution-based architectures such as the ResNet\cite{he2015deepresiduallearningimage} series and Transformer-based frameworks like Vision Transformer\cite{dosovitskiy2021imageworth16x16words}), learning paradigms (including single-task and multi-task modes), and supervision granularities (encompassing fully supervised and weakly supervised settings)\cite{xiao2024surveycamouflagedobjectdetection}. 
These deep learning-driven COD approaches enable the automatic learning of intricate discriminative features through neural networks. They can be categorized based on three core dimensions: network architectures, comprising linear, aggregative, branched, and hybrid types, with backbones typically involving convolution-based models (e.g., ResNet~\cite{he2015deepresiduallearningimage}) or Transformer-based frameworks (e.g., Vision Transformer~\cite{dosovitskiy2021imageworth16x16words}); learning paradigms, including single-task and multi-task modes ; and supervision granularities, encompassing fully supervised and weakly supervised settings~\cite{xiao2024surveycamouflagedobjectdetection, 9706783} developed a confidence-based COD framework with dynamic supervision, producing both camouflage masks and aleatoric uncertainty estimates, showing superior performance. \cite{10183371} proposed the Uncertainty-Edge Dual Guide model, which combines probabilistic uncertainty with deterministic edge information for accurate COD. Current research mainly addresses uncertainty in segmentation tasks, focusing on generating confidence maps for boundary distinction. However, in large-scale image data pre-training, the image space contains diverse object categories and complex structures. When the YOLO backbone, with relatively few parameters, learns spatial representations via joint embedding prediction, it tends to lack local representation consistency, and semantic structural features are prone to degenerating into average representations. To address this issue, we design regularization losses for feature invariance, variance, and covariance with dual similarity measurement, which balances the learning of similarity and independence between patch representations and alleviates the problem of representation collapse. Additionally, in the process of vision-text contrastive learning, camouflaged object detection (COD) relies solely on category-level text, lacking fine-grained semantic descriptions. This makes it difficult to form clear decision boundaries to distinguish between semantically highly similar backgrounds and camouflaged objects during contrastive learning. To tackle this challenge, we propose a visual feature selection and object description refinement strategy guided by object-specific descriptions; through contrastive learning between specific features and fine-grained text descriptions, the distance between camouflaged objects and backgrounds is effectively increased.

\section{Methodology}
%在本章，我们先介绍模型整体架构，再讲述我们提出的子描述主成分对比融合设计，以及特异性引导的空间弱对齐和动态聚焦策略，来构建一个强大的开放词汇伪装目标检测器。
In this section, we first present the overall architecture of our model, and then elaborate on the proposed \emph{sub-description principal component contrastive fusion} strategy, as well as the \emph{specificity-guided regional weak alignment and dynamic focusing scheme} strategy, which together enable the construction of a powerful open-vocabulary camouflaged object detector.

\subsection{Overall Framework}
\begin{figure*}
    \centering
    \includegraphics[width=\linewidth]{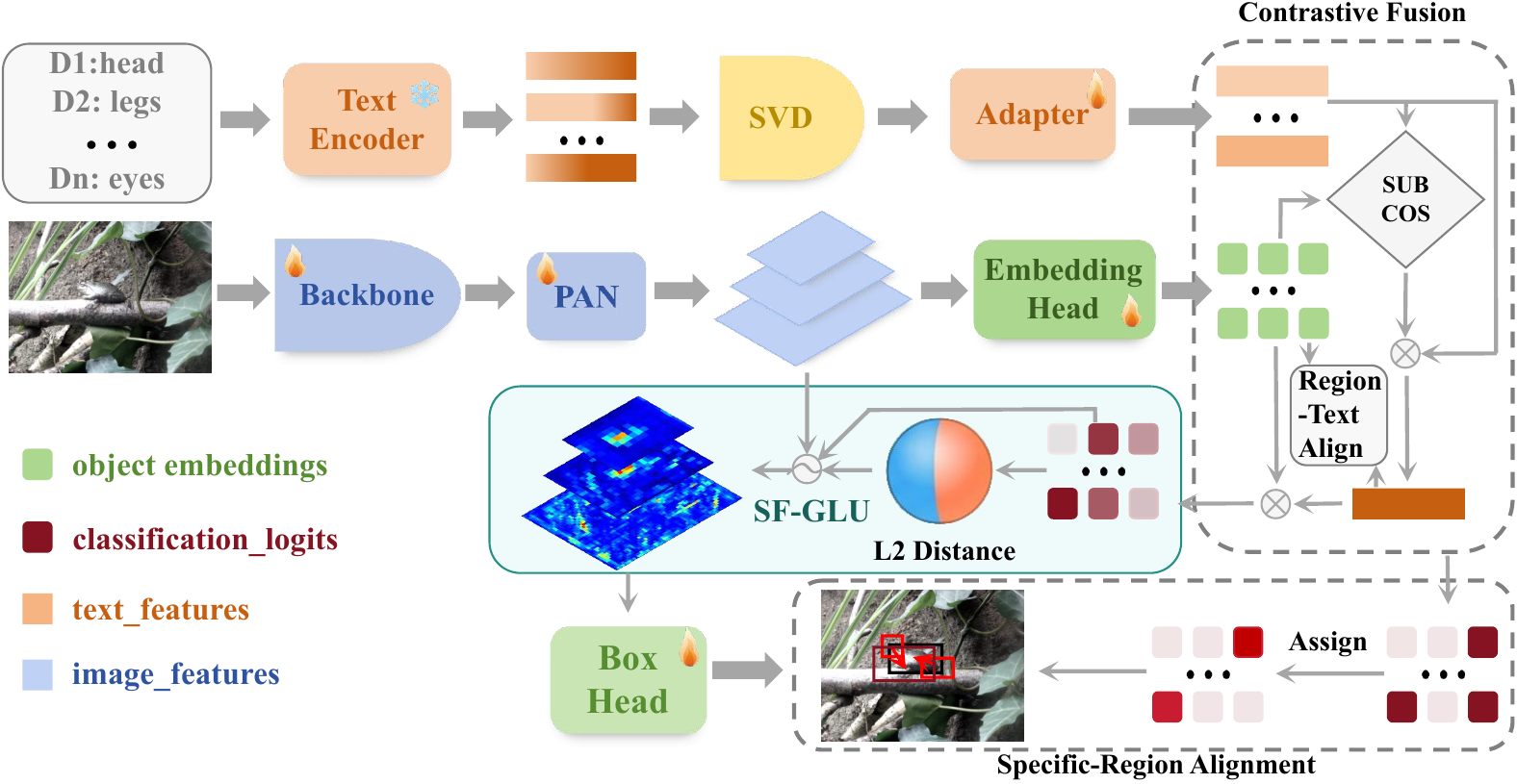}
    \caption{Overall architecture of the proposed specificity-driven open-vocabulary camouflaged object detector. Fine-grained textual sub-descriptions are encoded by a text encoder and decorrelated via SVD, then refined through an adapter and integrated with visual object embeddings in the contrastive fusion module. The image is processed by a lightweight YOLO‑style backbone and a PAN to extract multi‑scale features, which are transformed into object embeddings and fed in parallel into the SF‑GLU and the box head. 
}
    \label{fig:Overall Framework}
\end{figure*}

%模型架构如图所示，我们先将图像输入YOLO Backbone和PAN提取多尺度特征图，通过embedding head生成object embeddings。目标子描述经过文本编码得到子描述向量，融合后的文本向量与object embeddings进行区域文本匹配分类，增强后的视觉特征输入Bbox_head预测检测框。
As is depicted in Figure~\ref{fig:Overall Framework}, the input image is first fed into the YOLO backbone and PAN to extract multi-scale feature maps \(\mathbf{\hat{z}}^{(l)} \in \mathbb{R}^{H_l \times W_l \times C_l}\). These feature maps are then processed by the embedding head to produce a set of object embeddings \(\mathbf{v}_i \in \mathbb{R}^{d_v}\). In parallel, the object attribute sub-descriptions are encoded by a text encoder into sub-description vectors \(\{ \mathbf{t}_{c,k} \}_{k=1}^{K}\), which are further decorrelated via SVD and refined by an Adapter. The resulting sub-description principal components are integrated with \(\{ \mathbf{v}_i \}\) through our \emph{sub-description principal component contrastive fusion} module to obtain a fused textual representation \(\mathbf{t}_c^{\text{fused}}\) and discriminative region--text matching scores, enabling open-vocabulary recognition at the region level.

On top of this, a \emph{specific-region alignment} branch selects the object embedding that best matches \(\mathbf{t}_{c,k}\) and constrains its predicted box to weakly align with the ground-truth region. Meanwhile, the proposed \emph{SF-GLU} module uses both \(\mathbf{t}_c^{\text{fused}}\) and the region-wise similarity scores to generate a spatial gating signal that dynamically enhances the target-domain visual responses in \(\mathbf{\hat{z}}^{(l)}\). Finally, the enhanced visual features are passed to the Box Head to predict the detection bounding boxes \(\mathbf{\hat{b}}_i \in \mathbb{R}^4\).

\subsection{Sub-Description Principal Component Contrastive Fusion}

To more accurately distinguish camouflaged objects from the background, we employ fine-grained category attribute descriptions to refine the fused object description. However, because the object and the surroundings may share subtle local similarities, these category attribute descriptions can also contain ambiguous expressions. To mitigate such interference, we first perform SVD on the attribute descriptions to suppress redundant and noisy components.

% Formally, given the set of  sub-descriptions for category \(c\), denoted by \(\{ \mathbf{t}_{c,k} \}_{k=1}^{K}\), we obtain their decorrelated versions via
% \begin{equation}
%     \mathbf{\tilde{t}}_{c,k} = \operatorname{SVD}(\mathbf{t}_{c,k}), \quad k = 1, \dots, K.
% \end{equation}

Formally, given the set of  sub-descriptions for category \(c\), denoted by \(\{ \mathbf{t}_{c,k} \}_{k=1}^{K}\), we obtain their decorrelated and refined representations by applying SVD, followed by a three-layer MLP that serves as a text adapter: 
\begin{equation}
\mathbf{\tilde{t}}_{c,k} = \operatorname{Adapter}(\operatorname{SVD}(\mathbf{t}_{c,k})), \quad k = 1, \dots, K,
\end{equation}

%接着我们设计了一种对比式融合机制，我们计算patch特征vobj与子描述的相似度，跟同一子描述与全局patch特征vglobal的做差，得到重要性分数wk：

% wk=cos(vobj,tc,k)−cos(vglobal,tc,k).

Next, we design a contrastive fusion mechanism. Specifically, we compute the similarity between the  \(\mathbf{v}_{\text{i}}\) and each sub-description, and subtract from it the similarity between the same attribute description and the average object embedding \(\mathbf{v}_{\text{global}}\). This yields an importance score \(w_k\) for the \(k\)-th sub-description:
\begin{equation}
    w_k = \cos\bigl(\mathbf{v}_{\text{i}}, \mathbf{\tilde{t}}_{c,k}\bigr)
        - \cos\bigl(\mathbf{v}_{\text{global}}, \mathbf{\tilde{t}}_{c,k}\bigr),
\end{equation}

% 经softmax归一化后加权融合为类别向量tcfused=kαktc,k。该机制使模型自适应地选择判别性强的特异性子描述，显著提升语义判别性。

We then apply a softmax normalization to the importance score \(w_k\) to obtain the fusion weights \(\beta_k\), which is used to compute the fused object descriptions:  

\begin{equation}
\mathbf{t}_c^{\text{fused}} = \sum_{k=1}^{K} \beta_k \,\mathbf{\tilde{t}}_{c,k}, 
\end{equation}

This contrastive fusion mechanism enables the model to adaptively select highly discriminative, class-specific sub-descriptions, thereby substantially enhancing the semantic discriminability of the resulting representation.

\begin{figure*}
    \centering
    \includegraphics[width=\linewidth]{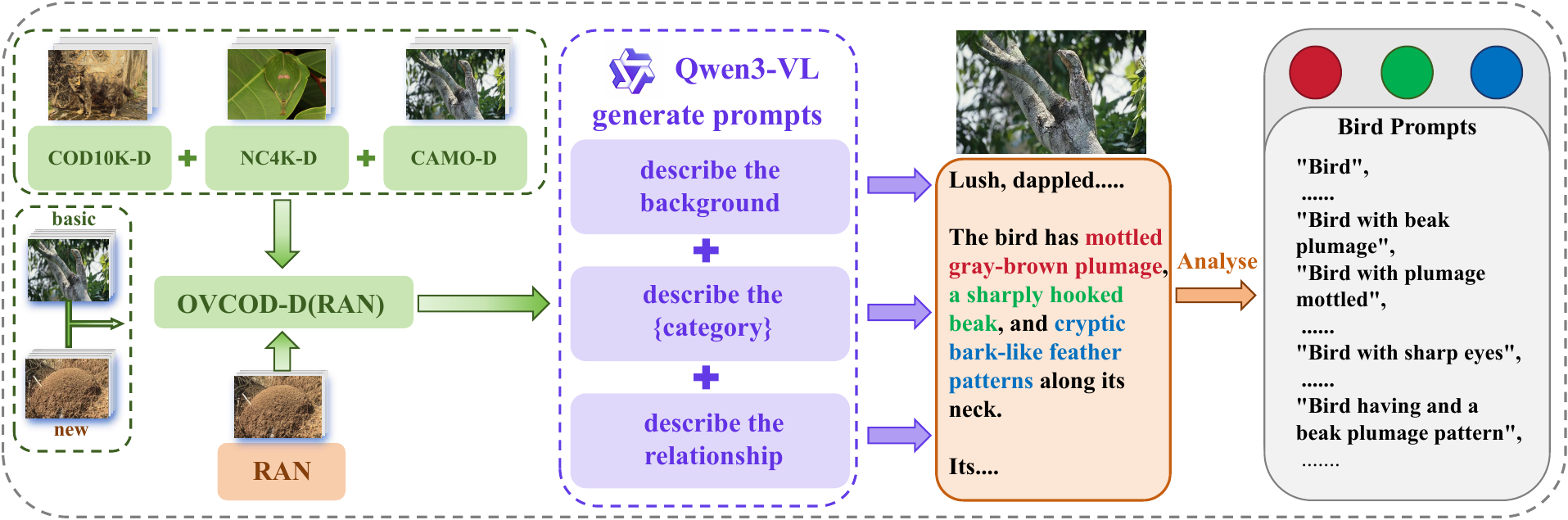}
    \caption{Construction pipeline of OVCOD-D dataset. We extend COD10K-D, NC4K-D, and clean CAMO-D with YOLO-style detection labels and an additional red imported fire ant nest subset, then reorganize them into 40 base and 47 novel classes. Qwen3-VL-Plus generates fine-grained image descriptions from which we derive a semantic prompt library for open-vocabulary camouflaged object detection.}
    \label{fig:benchmark}
\end{figure*}

% 特异性引导的空间弱对齐与动态聚焦策略
% 在弱视觉线索的情况下，尤其是伪装目标与周围环境高度相似，仅依靠目标整体语义不足以准确定位伪装目标，因此我们提出特异性区域预测，来辅助构建细粒度语义和空间的联系。受制于特异性区域标注的繁琐和不清晰性，我们采取弱监督方式，选择与特异性子描述高度匹配的区域，使其在空间上逼近目标区域内。定义修正器给object embedding分配的正样本为v_i+，同时v_{c,k}为区域与子描述相似度最高值对应的object embedding，两者同时满足才会将其对应的预测框\hat b_c作为对齐样本。

%约束其预测框bpred与真实标注bGT的覆盖比例：
% ℒac=1−|\hat b_{c,i}_c∩bGT||\hat b_{c,i}|
% 该损失显式鼓励特异性判别性强的区域在空间上对齐目标，使定位结果在视觉特征差异微弱时仍能收敛于目标区域
\subsection{Specificity-Guided Spatial Weak Alignment and Dynamic Focusing}

In the presence of weak visual cues, especially when camouflaged objects are highly similar to their surroundings, relying solely on global object semantics is insufficient for accurate localization. To alleviate this issue, we introduce \emph{specific-region alignment} to help establish a tighter connection between fine-grained semantics and spatial regions. However, due to the laborious and often ambiguous nature of annotating specificity regions, we adopt a weakly supervised strategy: we select regions that are highly consistent with the sub-descriptions and encourage them to spatially approach the ground-truth object area. We define the positive sample assigned by the refiner to an object embedding as \(\mathbf{v}_i^{+}\). Let $\mathbf{v}_{i_{c,k}}$ denote the object embedding whose corresponding region attains the highest similarity to the refined specific attribute description \(\mathbf{\tilde{t}}_{c,k}\) of category \(c\), i.e.,
\begin{equation}
    i_{c,k} = \arg\max_{i \in \{1,\dots,N\}}
    \operatorname{sim}\bigl( \mathbf{v}_i, \mathbf{\tilde{t}}_{c,k} \bigr),
\end{equation}

Only when the refiner-assigned positive sample coincides with this most similar embedding, do we treat the corresponding predicted box \(\mathbf{\hat{b}}_{c,i}\) as an alignment sample. We further constrain the coverage ratio between its predicted bounding box \(\mathbf{\hat{b}}_{c,i}\) and the ground-truth annotation \(\mathbf{b}_{\text{GT}}\) by introducing a coverage-based auxiliary loss:
\begin{equation}
    \mathcal{L}_{\text{ac}}
    = 1 - 
    \frac{\bigl| \mathbf{\hat{b}}_{c,i} \cap \mathbf{b}_{\text{GT}} \bigr|}
         {\bigl| \mathbf{\hat{b}}_{c,i} \bigr|},
\end{equation}
where \(|\cdot|\) denotes the area of a region. This loss explicitly encourages regions with strong specificity-based discriminative power to be spatially aligned with the true object, enabling the localization results to converge to the object area even when the visual differences between the object and the background are very subtle.

% 在特异性区域与目标区域对齐之上，为了让视觉分支能更聚焦于语义判别性强且区域相邻的patch特征来区分目标和背景，因此我们提出SF-GLU模块来增强目标域的视觉特征响应。首先计算文本区域相似度S_j，根据S最大值得到最有可能的目标patch i_o，并计算其他patch到目标patch的L2 距离d_{j, i_o}，将两者融合来生成门控信号：

%S=maxcσ(tcfused⊤vobj)， \hat z_enhanced=\hat z⊙(1+sigmoid(Conv(S) / r / d))

% 其中⊙为逐元素相乘，1+sigmoid(⋅)确保所有区域至少保留原始特征，我们希望语义判别性显著且空间距离相近区域获得额外增强。
On top of aligning the specificity regions with the object regions, we further encourage the visual branch to concentrate on patch features that are both semantically discriminative and spatially adjacent to the object region, so as to better distinguish the object from the background. To this end, we propose the SF-GLU module, which enhances the visual feature responses within the object domain.

We first compute the text–region similarity score \(S_j\) for each object embedding as
\begin{equation}
    S_j = \max_{c} \sigma\bigl( (t_c^{\text{fused}})^\top v_j \bigr),
    \quad j = 1, \dots, N,
\end{equation}

We then identify the most likely object patch index
\begin{equation}
    i_t = \arg\max_{j \in \{1,\dots,N\}} S_j,
\end{equation}

Next, we compute the spatial distance between each patch \(j\) and the object patch \(i_t\). Let 
\(\mathbf{p}_j \in \mathbb{R}^2\) denote the normalized spatial coordinate of the \(j\)-th patch; the 
L2 distance is given by
\begin{equation}
    d_{j,i_t} = \bigl\| \mathbf{p}_j - \mathbf{p}_{i_t} \bigr\|_2,
    \quad j = 1,\dots,N,
\end{equation}

We then fuse the similarity and distance information to generate the gating signal. In compact form, 
let \(\mathbf{S} = [S_1, \dots, S_N]\) and \(\mathbf{d} = [d_{1,i_t}, \dots, d_{N,i_t}]\); the SF-GLU 
operation is defined as
% \begin{equation}
%     \operatorname{SF\text{-}GLU}(\hat{z}, \mathbf{S}, \mathbf{d})
%     = \hat{z} \odot 
%     \Bigl( 
%         1 + \sigma\bigl( \operatorname{Conv}(\mathbf{S}  / (r \cdot \mathbf{d}) \bigr)
%     \Bigr).
% \end{equation}
\begin{equation}
    \operatorname{SF\text{-}GLU}(\hat{z}, \mathbf{S}, \mathbf{d})
    = \hat{z} \odot 
    \Bigl( 
        1 + \sigma\bigl( \operatorname{Conv}(\mathbf{S} / (r \cdot \mathbf{d})) \bigr) 
    \Bigr).
\end{equation}
% \begin{equation}
%     \operatorname{SF\text{-}GLU}(\hat{z}, \mathbf{S}, \mathbf{d})
%     = \hat{z} \odot 
%     \Biggl( 
%         1 + \sigma\left( \operatorname{Conv}\left( \frac{\mathbf{S}}{r \cdot \mathbf{d}} \right) \right) 
%     \Biggr).
% \end{equation}
where \(\odot\) denotes element-wise multiplication, and \(r > 0\) is a temperature parameter controlling the sharpness of the gating response. The term \(1 + \sigma(\cdot)\) ensures that all regions retain at least their original features, while 
regions that are both semantically discriminative and spatially close to the object receive additional enhancement, enabling the model to better focus on the camouflaged object domain.

%基于open-set设置上对比经典的开放词汇目标检测器在OVCOD-D数据集上的AP指标。我们将所有模型在基础类上微调，在基础类+新类上测试。
\begin{table*}[t]
  \centering
  \caption{Comparison with classical open-vocabulary object detectors on the OVCOD-D dataset is conducted under an open-set setting, where all models are fine-tuned on the base classes and evaluated on the union of base and novel classes. }
  \label{tab:open-set}
  \normalsize
  \setlength{\tabcolsep}{10pt}
  \begin{tabular}{lllcccccc}
    \toprule
    Method & Backbone & Params & Pre-train & AP & AP$_{50}$ & AP$_{75}$ & AP$_{m}$ & AP$_{l}$ \\
    \midrule
    GLIP-T~\cite{li2022grounded}            & Swin-T & 232M  & O365,GoldG & 22.6 & 31.9 & 28.2 & - & - \\
    GLIPv2-T~\cite{zhang2022glipv2unifyinglocalizationvisionlanguage}  & Swin-T & 232M  & O365,GoldG & 25.6 & 36.4 & 29.7 & - & -  \\
    \midrule
    Grounding DINO-T~\cite{liu2024grounding}           & Swin-T & 172M & O365,GoldG   & 34.8 & 43.9 & 37.7 & - & - \\
    \midrule
    YOLO-World-S~\cite{cheng2024yolo} & YOLOv8-S & 77M & O365,GoldG & 38.5 & 58.0 & 41.4 & 18.7 & 41.1\\
    YOLO-World-M~\cite{cheng2024yolo}                                                  & YOLOv8-M &  92M   & O365,GoldG  & 43.3 & 62.3 & 46.5 & 24.9 & 46.1 \\
    YOLO-World-L~\cite{cheng2024yolo}                                                  & YOLOv8-L &  110M   & O365,GoldG  &  45.7 & 63.2 & 48.9 & 22.9 & 48.4 \\
    \midrule
   YOLOE-S~\cite{wang2025yoloe}                     & YOLOv8-S &   78M   & O365,GoldG & 38.7 & 46.0 & 40.8 & - & - \\
    YOLOE-M~\cite{wang2025yoloe}                                                        & YOLOv8-M &    94M  & O365,GoldG & 39.9 & 47.7 & 42.7 & - & - \\
        % YOLOE-L                                                        & YOLOv8-L &    27M  & O365,GoldG & &  & & & \\
    \midrule
    DOSOD-S~\cite{he2025light}             & YOLOv8-S &  75M   & O365,GoldG & 44.8 & 67.4 & 46.8 & 22.4 & 47.2 \\
    DOSOD-M~\cite{he2025light}                                                      & YOLOv8-M &    90M       & O365,GoldG & 51.8 & 73.8 & 55.6 & 27.9 & 54.5 \\
    DOSOD-L~\cite{he2025light}                                                      & YOLOv8-L &    108M       & O365,GoldG & 53.4 & 73.1 & 56.2 & 26.4 & 56.3 \\
    \midrule
    \rowcolor{gray!12}
    SDDF-S     & YOLOv8-S &  76M          & O365,GoldG & \textbf{48.1} & 70.7 & 50.3 & 25.9 & 50.7 \\
    \rowcolor{gray!12}
    SDDF-M     & YOLOv8-M &  91M     & O365,GoldG & \textbf{54.3} & 75.1 & 57.5 & 30.3 & 57.0 \\
    \rowcolor{gray!12}
    SDDF-L     & YOLOv8-L &  109M        & O365,GoldG & \textbf{56.4} & 76.4 & 60.7 & 34.4 & 59.0 \\
    \bottomrule
  \end{tabular}
  \vspace{4pt}
\end{table*}

\section{Experiments}

\subsection{Benchmarking Open-Vocabulary Camouflaged Object Detection}
%在搭建OVCOD基准的过程中，我们基于伪装目标检测的图像数据，融合多模态大模型生成文本提示，将其拓展为开放词汇形式的数据集。接着我们将经典的开放词汇目标检测器在OVCOD数据集上进行微调，得到开放词汇伪装目标检测器的baseline。
\noindent 
In constructing the OVCOD-D benchmark, we build upon existing camouflaged object detection image datasets and integrate text prompts generated by large multimodal models, thereby expanding them into an open-vocabulary format. Subsequently, we fine-tune classical open-vocabulary object detectors on the OVCOD-D dataset to establish the baseline for open-vocabulary camouflaged object detection.

\noindent \textbf{Dataset Construction.} Existing COD datasets focus on segmentation and lack detection-ready boxes and category labels~\cite{xiao2024surveycamouflagedobjectdetection}. As is described in Fig~\ref{fig:benchmark}, to enable open-vocabulary COD, we extend COD10K-D, NC4K-D, and CAMO-D~\cite{xin2025toward}, derive consistent bounding boxes and categories from masks, and standardize annotations to a YOLO-compatible TXT format. We curate CAMO-D by removing non-camouflage images and add a self-constructed red fire ant nest set due to its agricultural and health significance. We restructure categories and splits, yielding 40 base classes and 47 novel classes (unseen during training). To mitigate object–background semantic similarity, we employ Qwen3-vl-plus\cite{yang2025qwen3technicalreport} to generate fine-grained per-image descriptions capturing background cues, object specificity, and spatial relations; from these, we distill category-level physical-feature descriptors and their frequencies into a semantic text repository. Overall, the dataset remedies detection-annotation and open-vocabulary gaps and strengthens vision–language linkage, providing high-quality, multidimensional data for training and evaluation in OVCOD. 

In summary, the dataset constructed in this work not only remedies the shortcomings of existing COD datasets in detection annotations and open-vocabulary compatibility but also strengthens the linkage between vision and language through fine-grained semantic descriptions and a semantic text repository. It provides high-quality, multidimensional data support for training and evaluating models in open-vocabulary camouflaged object detection.

% 现有COD数据集以分割为主，缺少检测所需的框与类别标注。为支持开放词汇COD，我们扩展 COD10K-D、NC4K-D 与 CAMO-D~\cite{xin2025toward}，在原始掩码基础上生成一致的框与类别，并将注释统一为 YOLO 兼容的 TXT 格式。我们清理 CAMO-D，移除非伪装图像，并新增自建的红火蚁巢数据，考虑其对农业与健康的危害。在合并数据上重构类别与划分，形成 40 个基类与 47 个新类（训练不可见）。为缓解目标与背景的语义相似，使用 Qwen3-vl-plus 生成每图细粒度文本，刻画背景线索、目标物理属性与空间关系；进一步提炼各类别的辨识性物理特征并统计频次，构建语义文本库。总体而言，该数据集弥补检测标注与开放词汇兼容的不足，强化视—语联动，为开放词汇伪装目标检测的训练与评估提供高质量、维度丰富的数据支撑。

%benchmark第五版    修改者 郑纬华  2025/11/10

\noindent \textbf{Model Variants.} Our experiments evaluate several popular zero-shot detection models on LVIS, employing various architectural frameworks such as Transformer and CNN. Regarding the backbones, GLIP-T~\cite{li2022grounded}, GLIPv2-T~\cite{zhang2022glipv2unifyinglocalizationvisionlanguage}, and Grounding DINO-T~\cite{liu2024grounding}  utilize a Swin-Transformer architecture. Meanwhile, YOLO-World, YOLOE~\cite{wang2025yoloe}, and DOSOD~\cite{he2025light} are based on YOLOv8, a CNN-based framework with residual structures. All these models are pre-trained on O365 and GoldG datasets. In terms of parameters, models like GLIP-T have 232M params, Grounding DINO-T has 172M, while YOLO-World-S has only 13M — showcasing a wide range of parameter scales. We use their official implementations and pre-trained weights for the experimental setup.

\noindent \textbf{Evaluation Metrics.}
%跟随经典的开放词汇检测benchmark，我们使用AP来衡量检测器的表现。由于这些benchmark用于通用目标检测，所以他们按照目标的样本量比例分为稀有、常见、高频目标，性能分别对应AP_r、AP_c、AP_f。而伪装目标本身是一种特殊分级，大多都是稀有品种，因此按样本量比例分类不合理，所以我们报告不同IoU和不同目标尺寸的AP。
Following the conventions established by classic open-vocabulary detection benchmarks~\cite{zareian2021open, gu2021open, gupta2019lvis}, we evaluate the performance of our detector using the Average Precision (AP) metric. Since these benchmarks are designed for general object detection, they categorize objects into \textit{rare}, \textit{common}, and \textit{frequent} classes based on the number of available samples, and report the corresponding metrics as $\mathrm{AP}_r$, $\mathrm{AP}_c$, and $\mathrm{AP}_f$, respectively. However, most camouflaged instances are inherently rare, rendering classification based on sample frequency inappropriate. Therefore, instead of reporting AP by category frequency, we present results in terms of AP at different Intersection-over-Union (IoU) thresholds and across various object scales.

\subsection{Comparisons on Open-Vocabulary Camouflaged Object Detection}
%为了衡量检测器在开放词汇伪装目标检测benckmark上的零样本检测能力，我们将经典的开放词汇目标检测器在基础类上微调，在基础类和新类上进行测试。考虑到经典的开放词汇目标检测器是通过单个目标描述来检测目标，我们在baseline模型上加入同一种目标子描述向量的融合方式，来保证文本信息的公平性。除此之外，为了维持实验的公平性，baseline模型和提出的方法保持学习率，batch_size，随机种子等实验设置一致。需要注意的是，由于伪装目标数据规模的限制，导致其难以模仿通用目标形式来划分大规模新类进行零样本检测，所以模型性能的量级不一致。对比同样使用轻量化yolo backbone和PAN的开放词汇检测器，在最小的模型上我们的方法比表现最好的yoloe提高8.9%的AP。相较于大型开放词汇检测器，性能最佳的Grounding DINO-T也与我们的方法存在差距，足以证明SDDF在未知伪装类别上强大的零样本检测能力。随着模型参数上涨，SDDF也相较其他开放词汇检测器出现更大幅度的性能提升,最大模型比最小模型的AP高出8.8。对于不同大小的伪装目标，SDDF对中型伪装目标检测的能力提升较为显著，取得11.0的AP增值。
%分析glip和gdino性能低的原因
%值得注意的是，GLIP 和 Grounding DINO 等模型在我们的数据集上性能表现不佳。这源于其 Token 级的短语定位机制与密集、多属性子描述的 Token 化过程之间存在的不匹配。通过 BERT 分词器处理复杂的子描述提示，会使得离散的语义概念碎片化，并导致局部上下文语义模糊。再加上严格的序列长度限制，这彻底破坏了其深层跨模态注意力机制。
\noindent To evaluate the open-vocabulary detection capability of the detector on the proposed OVCOD-D, we fine-tune classical open-vocabulary object detectors on the base classes and then test them on both base and novel classes. 

Considering that classical open-vocabulary detectors typically rely on a single textual description per object category, we extend the baseline models by incorporating the same sub-description fusion mechanism detailed in the appendix, so as to ensure fairness with respect to the textual information. It is worth noting that, due to the limited scale of camouflaged object data, it is difficult to construct large-scale novel categories in a manner analogous to generic object detection benchmarks for zero-shot evaluation~\cite{gupta2019lvis}. As a consequence, the absolute magnitude of the reported performance is not directly comparable to that of conventional zero-shot object detection settings.

As shown in Table~\ref{tab:open-set}, compared with other open-vocabulary detectors that also adopt a lightweight YOLO backbone and PAN, our method outperforms the best-performing DOSOD variant by 3.3 AP on the smallest model configuration. It is noteworthy that existing grounding models, such as GLIP and Grounding DINO, exhibit suboptimal performance on our dataset. This degradation primarily stems from a fundamental misalignment between their token-level phrase-grounding mechanisms and the tokenization process of dense, multi-attribute sub-descriptions. Specifically, processing complex sub-description prompts via the BERT~\cite{devlin2019bert} tokenizer tends to fragment discrete semantic concepts and induce local contextual ambiguity. Such issues, further exacerbated by stringent sequence length constraints, ultimately compromise the efficacy of their deep cross-modal attention mechanisms. 
As the model size increases, SDDF exhibits an even more pronounced performance gain over other open-vocabulary detectors, with the largest model achieving an AP that is 8.3 points higher than that of the smallest one. In terms of object scale, SDDF yields particularly notable improvements for medium-sized camouflaged targets, achieving an AP increase of 8.5.
\begin{table}[t] 
  \centering
  \caption{Comparisons under the closed-set setting are conducted to evaluate the performance of classical object detectors and open-vocabulary object detectors on the OVCOD-D dataset. The FPS is measured on a single NVIDIA RTX 4090 using the raw trained checkpoints without TensorRT, re-parameterization, or other lightweight optimizations.
}
  \label{tab:coco_yolo_comparison}
  \resizebox{\linewidth}{!}{  
    \begin{tabular}{l c c c c c}
      \toprule
      Method & Pre-train & AP & AP$_{50}$ & AP$_{75}$ & FPS \\
      \midrule
      \multicolumn{6}{l}{\scriptsize Training from scratch.} \\ 
      YOLOv8-S~\cite{10533619} & $\times$ & 18.6 & 26.5 & 19.0 & - \\
      YOLOv8-M~\cite{10533619} & $\times$ & 21.7 & 31.8 & 22.6 & - \\
      YOLOv8-L~\cite{10533619} & $\times$ & 23.7 & 32.0 & 23.4 & - \\
      \midrule
      \multicolumn{6}{l}{\scriptsize Fine-tuned w/ DuVL-PAN and sub-descriptions} \\
      YOLO-World-S~\cite{cheng2024yolo} & O+G & 39.3 & 57.3 & 41.3 & 35 \\
      YOLO-World-M~\cite{cheng2024yolo} & O+G & 45.0 & 65.2 & 47.6 & 33 \\
      YOLO-World-L~\cite{cheng2024yolo} & O+G & 46.9 & 66.9 & 49.4 & 29 \\
      % \midrule
      % \multicolumn{6}{l}{\scriptsize Fine-tuned w/ sub-descriptions} \\
      % YOLOE-S~\cite{wang2025yoloe} & O+G & 40.1 & 47.3 & 41.2 & 47 \\
      % YOLOE-M~\cite{wang2025yoloe} & O+G & 42.5 & 49.1 & 43.1 & 39 \\
      % YOLOE-L & O+G &  &  &  & - \\
      \midrule
      \multicolumn{6}{l}{\scriptsize Fine-tuned w/ sub-descriptions} \\
      DOSOD-S~\cite{he2025light} & O+G & 44.9 & 67.2 & 48.7 & 44 \\
      DOSOD-M~\cite{he2025light} & O+G & 52.9 & 74.0 & 55.1 & 42 \\
      DOSOD-L~\cite{he2025light} & O+G & 53.6 & 72.9 & 56.8 & 36 \\
      \midrule
      \multicolumn{6}{l}{\scriptsize Fine-tuned w/ sub-descriptions} \\
      \rowcolor{gray!12}
      SDDF-S & O+G & \textbf{48.2} & 70.1 & 51.7 & 39 \\
      \rowcolor{gray!12}
      SDDF-M & O+G & \textbf{54.8} & 73.6 & 58.1 & 36 \\
      \rowcolor{gray!12}
      SDDF-L & O+G & \textbf{59.4} & 80.3 & 63.4 & 33 \\
      \bottomrule
    \end{tabular}
  }
  
  \vspace{2pt} 
\end{table}

%在伪装目标检测这一任务上，传统的方法的检测头大都用于实例分割~\cite{}，即给出伪装目标的细粒度掩膜。因此在目标分类定位的任务设置上，我们选择经典的YOLO系列模型，包括基于其设计的开放词汇检测器，在所有类别上进行训练和测试。实验设置上，为了维持模型融合细粒度文本描述的潜质，我们在闭集伪装检测上也使用目标子描述，同时将batchsize设置成32、统一学习率来保证实验公平。根据Table2的数据，SDDF在最小模型上取得48.2的AP值，对比同等baseline中性能最好的方法高了 AP。随着模型参数变大，SDDF-l比SDDF-S在AP指标上提升了11.2，同时我们发现在高质量的预测框中（IoU>0.75），我们的模型性能增幅最大，进一步证实其非凡的伪装目标检测能力。在推理时延上，模型也达到 FPS，有助于其在端侧部署和应用。
\subsection{Comparisons on Camouflaged Object Detection}
For the task of camouflaged object detection, traditional methods mostly adopt detection heads designed for instance segmentation~\cite{xiao2024surveycamouflagedobjectdetection}, which output fine-grained masks of camouflaged objects. Therefore, under a task setting focused on object classification and localization with bounding boxes, we choose the classical YOLO-series models, including open-vocabulary detectors derived from them, and train and evaluate all categories jointly.

In terms of experimental setup, to preserve the model’s ability to exploit fine-grained textual descriptions, we also employ target sub-descriptions in the closed-set camouflaged detection experiments. Moreover, we fix the batch size to 32 and use a unified learning rate across all methods to ensure a fair comparison. According to the results in Table ~\ref{tab:coco_yolo_comparison} , SDDF achieves an AP of 48.2 with the smallest model configuration, outperforming the strongest baseline under the same model size.

As the model capacity increases, SDDF-L improves AP by 11.2 points over SDDF-S. We further observe that the performance gain is most pronounced for high-quality predictions (IoU $> 0.75$), which provides additional evidence of its exceptional capability in camouflaged object detection. In terms of inference latency, the model also operates at real-time frame rates, facilitating deployment and application on edge devices.

% Some key observation here, e.g. center in action
\section{Ablation Study}\label{Sec:ablated_center}

%我们针对模型架构中子描述文本的融合策略进行了消融实验，旨在探究对比融合策略的有效性。在对特异性主成分进行融合时，考虑到正交化融合策略，可以在数学上确保不同子描述（如颜色、纹理、形状等）在语义空间中投射到互不重叠的方向，有利于保留特异语义的多元性。因此我们设计多组对照实验，分为拼接融合、正交融合策略和正交对比融合策略。根据Table 3的实验结果，我们提出的对比融合策略AP性能高达47.5，猜测模型利用子描述和目标与背景的相似度差异，有利于判别子描述的重要性来融合出更能反映伪装目标和背景区分度的目标描述向量。
\subsection{Ablation study on contrastive fusion module}
We conduct ablation studies on the sub-description text fusion strategy to investigate the effectiveness of the proposed contrastive fusion mechanism. When fusing the specificity-aware principal components, the orthogonal fusion strategy is particularly appealing, as it can mathematically ensure that different sub-descriptions (e.g., color, texture, shape) are projected onto mutually non-overlapping directions in the semantic space, thereby preserving the diversity of specific semantics. 
To this end, we design several controlled experiments, comparing three variants:  (i) orthogonal fusion, and (ii) the proposed orthogonal contrastive fusion strategy. As shown in Table~\ref{tab:text_fusion_ablation}, our contrastive fusion strategy achieves an AP of 47.5, the best among all variants. We conjecture that the model benefits from explicitly leveraging the similarity differences between sub-descriptions and target versus background regions, which helps it assess the relative importance of each sub-description that better reflects the discriminability between camouflaged objects and the background.

\begin{table}[t]
\centering
\caption{Ablation study on sub-description fusion strategies for open-vocabulary camouflaged object detection on OVCOD-D dataset. The coverage auxiliary loss weight is uniformly set to 5, and SF-GLU is disabled for all variants to ensure fair comparison.}
\label{tab:text_fusion_ablation}
\begin{tabular}{lccc}
\hline
Fusion Method & AP & AP\textsubscript{50} & AP\textsubscript{75} \\
\hline
Orthogonal+Contrastive  & 46.5 & 70.1 & 48.9 \\
Contrastive   & \textbf{47.5} & \textbf{70.6} & \textbf{49.6 }\\
Orthogonal  &44.0  &66.4  &45.6  \\
\hline
\end{tabular}
\end{table}

\subsection{Ablation study on SF-GLU module}
%本文提出的SF-GLU模块在视觉特征动态增强起到关键作用，为了研究SF-GLU模块的其他变体，在文本区域相似度作用下对视觉特征增强的影响。我们设计了几个对照实验：不使用GLU、GLU模块、SwiGLU模块，这几种模块的作用本质上都是在改变视觉特征分布的锐化程度。根据表格的实验，搭载SF-GLU模块的模型在开集测试中达到47.6 AP, 猜测SF-GLU模块能通过空间距离锐化视觉特征分布，促使网络动态聚焦于目标域来区分开伪装目标和背景。此外，SF-GLU模块能让模型产生高质量的精准预测框，在IoU大于0.75的AP指标达到50.5。
% 此外，SF-GLU模块增强了模型对目标的高精度定位能力。此外，相较其他模块，SF-GLU能够更有效产生精准的伪装目标定位框（AP_75高达50.5）。
% Moreover, the proposed SF-GLU module improves the model’s ability to localize objects with high precision. Under the evaluation criterion of IoU ≥ 0.75, SF-GLU achieves an AP of 50.5, outperforming all other variants and indicating that it effectively enhances the generation of fine-grained and tightly aligned bounding boxes.

To investigate the effects of alternative variants of the SF-GLU module under text–region similarity guidance, we design several ablation experiments, including variants that remove the GLU module entirely or replace it with the standard GLU or SwiGLU modules. These modules fundamentally operate by modulating the sharpness of the visual feature distribution.

According to the experimental results presented in the Table~\ref{tab:glu_ablation}, the model equipped with the SF-GLU module achieves an AP of 48.1 in open-set evaluation. We hypothesize that the SF-GLU module sharpens the visual feature distribution through spatial-distance–aware modulation, enabling the network to dynamically focus on the target regions and effectively distinguish camouflaged objects from the background. In addition, compared with other module variants, SF-GLU is able to generate more accurate localization boxes for camouflaged objects, achieving an $\mathrm{AP}_{75}$ of 50.3.

\begin{table}[t]
% \caption{Ablations on SF-GLU module. }
\caption{Ablation study on SF-GLU module for open-vocabulary camouflaged object detection on the OVCOD-D dataset. All variants adopt the contrastive fusion framework with coverage auxiliary loss set to 5. }
\label{tab:glu_ablation}
\centering
\begin{tabular}{lccc}
\hline
Dynamic Focusing Module     & AP            & AP\textsubscript{50}     & AP\textsubscript{75}     \\
\hline
w/o GLU & 47.5          & 70.6          & 49.6          \\
GLU              & 46.7          & 69.7          & 49.8          \\
SwiGLU            & 47.1          & 70.3          & 49.2          \\
SF-GLU            &    \textbf{48.1}           &      \textbf{70.7}     &    \textbf{50.3}               \\
\hline
\end{tabular}
\end{table}

\subsection{Ablation study on coverage-based auxiliary loss}
% 为了探究覆盖率辅助损失对伪装目标检测性能的影响，我们对覆盖率辅助损失的权重参数λ进行了消融实验，目的在于确定最佳的覆盖率损失贡献程度，以平衡分类准确性和定位精度。如表3所示，我们测试了λ在{0,2,5,7,10}的取值，其中λ=0表示完全关闭覆盖率辅助损失。实验结果表明，当 $\lambda$ 从 0 增至 5 时，虽然 $AP_{50}$下降了 0.4，但 $AP_{75}$依然保持稳定并微增。 这说明覆盖率损失，剔除了预测框中的背景成分，使定位结果更加紧密。当 $\lambda \ge 7$ 时，AP 大幅下降。 这种反应反向证明了该损失施加的是一种强物理约束，当约束过强时，它会为了几何对齐而扭曲特征学习，从而验证了其空间限制的有效性。最终我们选择λ=5为最终的权重。

To assess the impact of the coverage-aware auxiliary loss, we performed an ablation study on its weighting parameter $\lambda$, seeking the optimal balance to yield high-quality localization results. As shown in Table~\ref{tab:coverage_weight_ablation}, we evaluated $\lambda$ with values in $\{0, 2, 5, 7, 10\}$, where $\lambda = 0$ denotes the complete removal of the coverage-aware auxiliary loss. 
% Experimental results indicate that setting $\lambda = 5$ yields a significant improvement in localization accuracy, demonstrating that the coverage loss effectively guides the model to focus on semantically relevant regions. When $\lambda \geq 7$, the AP decreases, suggesting that an excessively strong coverage constraint overemphasizes spatial alignment, thereby impairing classification capability. Conversely, when $\lambda = 2$, the weighting is too weak to provide sufficient guidance from the coverage loss, resulting in no AP improvement. 
Experimental results demonstrate that when $\lambda$ increases from 0 to 5, although $AP_{50}$ decreases by 0.4, $AP_{75}$ remains stable with a slight improvement. This indicates that the coverage loss effectively eliminates background components within the predicted boxes, leading to more compact localization results. However, when $\lambda \ge 7$, the overall AP drops significantly. We hypothesize that excessive constraints may distort feature learning in favor of geometric alignment, potentially biasing the model to predict bounding boxes closer to the ground- truth by misinterpreting features from non-target regions.
Therefore, we ultimately adopt $\lambda = 5$ as the final weighting parameter.

\begin{table}[t]
\centering
\caption{Ablation study on different weights of coverage-based auxiliary loss for open-vocabulary camouflaged object detection on the OVCOD-D dataset. All variants adopt the same contrastive fusion framework without using the SF-GLU module.}
\label{tab:coverage_weight_ablation}
\begin{tabular}{cccc}
\toprule
Weight for auxiliary loss & AP & AP\textsubscript{50} & AP\textsubscript{75} \\
\midrule
0 & 47.3 & \textbf{71.0} & 49.5 \\
2 & 47.2 & 70.5 & 48.7 \\
5 & \textbf{47.5} & 70.6 & \textbf{49.6} \\
7 & 46.3 & 69.3 & 49.0 \\
10 & 46.6 & 69.7 & 49.0 \\
\bottomrule
\end{tabular}
\end{table}

% \begin{table}[t] 
%   \centering
%   \caption{Comparison of Detection Performance on Benchmark Datasets}
%   \label{tab:detection_comparison}
%   \resizebox{\linewidth}{!}{ 
%     \scriptsize  
%     \begin{tabular}{l c c c c}
%       \toprule
%       Method & AP & AP$_r$ & AP$_c$ & AP$_f$ \\
%       \midrule
%       ViLD {[}13{]} & 27.8 & 16.7 & 26.5 & 34.2 \\
%       RegionCLIP {[}62{]} & 28.2 & 17.1 & - & - \\
%       Detic {[}63{]} & 26.8 & 17.8 & - & - \\
%       FVLM {[}22{]} & 24.2 & 18.6 & - & - \\
%       \midrule
%       DetPro {[}8{]} & 28.4 & 20.8 & 27.8 & 32.4 \\
%       BARON {[}53{]} & 29.5 & 23.2 & 29.3 & 32.5 \\
%       \midrule
%       YOLOv8-S & 19.4 & 7.4 & 17.4 & 27.0 \\
%       YOLOv8-M & 23.1 & 8.4 & 21.3 & 31.5 \\
%       YOLOv8-L & 26.9 & 10.2 & 25.4 & 35.8 \\
%       \midrule
%       YOLO-World-S & 23.9 & 12.8 & 20.4 & 32.7 \\
%       YOLO-World-M & 28.8 & 15.9 & 24.6 & 39.0 \\
%       YOLO-World-L & 34.1 & 20.4 & 31.1 & 43.5 \\
%       \bottomrule
%     \end{tabular}
%   }
  
%   \vspace{2pt} 
% \end{table}

\section{Conclusion}
In this work, we address the largely overlooked challenge of open-vocabulary camouflaged object detection, where significant visual similarity between targets and their surroundings often hinders reliable localization and recognition. To facilitate research in this area, we construct the OVCOD-D benchmark by incorporating fine-grained textual descriptions into carefully curated camouflaged object images. Building upon detectors pre-trained on large-scale object detection datasets, we further introduce two key contributions: (1) a sub-description principal component contrastive fusion strategy that suppresses noisy or overly decorative textual modifiers, and (2) a specificity-guided regional weak alignment and dynamic focusing mechanism that enhances the model’s ability to distinguish camouflaged objects from background regions.

Comprehensive experiments under the open-set evaluation setting demonstrate that our method achieves an AP of 56.4 on the OVCOD-D benchmark, substantially improving the detector’s discrimination capability in complex camouflage scenarios. Overall, this study establishes a strong foundation for future research on open-vocabulary camouflaged object detection.

{
    \small
    \bibliographystyle{ieeenat_fullname}
    % \bibliography{main,robust,acmart}
    \bibliography{main}

}


%rebuttal
% Import additional packages in the preamble file, before hyperref

\input{preamble}
% If you comment hyperref and then uncomment it, you should delete
% egpaper.aux before re-running latex.  (Or just hit 'q' on the first latex
% run, let it finish, and you should be clear).
% \definecolor{cvprblue}{rgb}{0.21,0.49,0.74}
% \usepackage[pagebackref,breaklinks,colorlinks,allcolors=cvprblue]{hyperref}
% \usepackage[capitalize]{cleveref}
% \crefname{section}{Sec.}{Secs.}
% If you wish to avoid re-using figure, table, and equation numbers from
% the main paper, please uncomment the following and change the numbers
% appropriately.
%\setcounter{figure}{2}
%\setcounter{table}{1}
%\setcounter{equation}{2}

% If you wish to avoid re-using reference numbers from the main paper,
% please uncomment the following and change the counter value to the
% number of references you have in the main paper (here, 100).
%\makeatletter
%\apptocmd{\thebibliography}{\global\c@NAT@ctr 100\relax}{}{}
%\makeatother

%%%%%%%%% PAPER ID  - PLEASE UPDATE
\def\paperID{35654} % *** Enter the Paper ID here
\def\confName{CVPR}
\def\confYear{2026}

% \begin{document}
\definecolor{MyDarkGreen}{RGB}{40,120,50}
\definecolor{MyDarkBlue}{RGB}{0,0,205}
\definecolor{MyPurple}{RGB}{117,43,168}
\definecolor{MyPink}{RGB}{231,72,151}

\vspace{24pt}

% \clearpage
\appendix

% 重置图、表、公式的计数器，并加上 "A" 前缀
\setcounter{section}{0}
\setcounter{figure}{0}
\renewcommand{\thefigure}{A\arabic{figure}}
\setcounter{table}{0}
\renewcommand{\thetable}{A\arabic{table}}
\setcounter{equation}{0}
\renewcommand{\theequation}{A\arabic{equation}}

\renewcommand{\theHequation}{A\arabic{equation}}

\noindent\textbf{\Large Appendix}
\vspace{10pt}

In this appendix, we provide additional implementation details and extensive experimental results to further elucidate our proposed method. The supplementary material is organized as follows:
\begin{itemize}
    \item In Section~\ref{sec:dataset_details}, we present a comprehensive description of the dataset construction process, including details on data collection, annotation protocols, and novel class overlap analysis.
    \item In Section~\ref{sec:exp_details}, we delineate the complete experimental pipeline, baselines' text pipeline, and provide a full listing of hyperparameter configurations and numerical stability designs.
    \item In Section~\ref{sec:add_quant}, we provide additional quantitative results, including comparison with state-of-the-art COD methods and analysis across specific camouflage difficulty levels.
    \item In Section~\ref{sec:ablation_ext}, we conduct extended ablation studies examining the influence of SVD configurations and the SF-GLU convolutional layers.
    \item In Section~\ref{sec:qualitative}, we perform detailed qualitative comparisons through visualization of detection bounding boxes and feature maps.
\end{itemize} 

% ==========================================
% Section A: Dataset Information
% ==========================================
\section{Dataset Information and Analysis}
\label{sec:dataset_details}

\subsection{OVCOD-D Construction and Quality Assurance}
The newly constructed OVCOD-D benchmark comprises 6{,}469 training samples and 3{,}957 test samples, into which 83 high-quality images of red imported fire ant (RIFA) nests have been incorporated. All RIFA nest images were collected in natural field environments, thereby addressing a gap in existing datasets concerning camouflaged insect-nest targets. 

For a reproducible construction pipeline, we established a standardized three-stage pipeline: (a) Instance masks were converted to axis-aligned YOLO-format boxes. Coordinates were normalized with strict boundary clamping to ensure validity. (b) We filtered COD10K, retaining only the subset of camouflaged ones. (c) Label spaces were unified via case-insensitive matching and modifier removal, yielding an 87-category vocabulary.

For generating fine-grained sub-descriptions of target objects, we employed the following prompt:
\begin{quote}
The image contains a \{display\_name\}. \textbf{YOU MUST GENERATE EXACTLY THREE SEPARATE SENTENCES.} \\
\textbf{FIRST SENTENCE:} Describe the background in \textbf{ONE VERY CONCISE English sentence} containing \textbf{AT LEAST THREE ADJECTIVES}, without mentioning the \{display\_name\}. \\
\textbf{SECOND SENTENCE:} Only describe the \{display\_name\} in \textbf{ANOTHER VERY CONCISE English sentence} with \textbf{AT LEAST THREE PHYSICAL FEATURES} that distinguish it from its surroundings. \textbf{DO NOT MENTION} “background,” “environment,” “surroundings,” or any context-related words. \\
\textbf{THIRD SENTENCE:} Describe the physical relationship between the \{display\_name\} and its background in \textbf{ONE VERY CONCISE English sentence}.
\end{quote}

Subsequently, we obtain fine-grained object descriptions for each category and process them accordingly. Through a series of text preprocessing steps, including the removal of punctuation, conversion to lowercase, tokenization, and stop-word filtering. We extract semantically meaningful specificity-describing terms. For each species, we then construct a term-frequency statistics profile and retain the top 30 highest-frequency terms to form a species-specific vocabulary repository. We generate over 20 fine-grained sub-descriptions per class (4–8 valid tokens on average), with CLIP-native tokenization and the CLIP text encoder for text encoding. Each sub-description incorporates both the original scientific name of the species and its plural form. Furthermore, we select the top six pairs of high-frequency terms and assemble them into structured phrases, thereby ensuring that at least 20 semantically rich descriptive phrases are produced for each species.

For Text Quality Assurance, as shown in Table~\ref{tab:human_eval_text_desc}, five sub-descriptions are randomly sampled per category (or all if fewer than five are available). Positive and negative samples are identified based on three pre-defined metrics to calculate the precision rate. A category is deemed ``qualified'' if over 50\% of its sampled sub-descriptions are identified as positive.

\begin{table}[h]
\centering
\caption{Human evaluation of generated textual descriptions.}
\tabcolsep=4pt
\resizebox{0.95\columnwidth}{!}{
% \scriptsize
% \small
\footnotesize
\begin{tabular}{c|ccc|cc}
\hline
Metric & Accuracy & Relevance & Hallucination-free & Qualified category \\
\hline
Rate (\%) & 98.15 & 99.77 & 92.61  & 100.00 \\
\hline
\end{tabular}
}
\label{tab:human_eval_text_desc}
\end{table}

\subsection{Novel Class Overlap with Pretraining Data}
26 out of 47 novel classes have no overlap with pre-training data. Nevertheless, the domain gap induced by high camouflage prevents model from relying on pre-learned features. 

% ==========================================
% Section B: Implementation Setup
% ==========================================
\section{Implementation Setup}
\label{sec:exp_details}

We conduct all experiments using PyTorch 2.1.2 with CUDA 11.8 on 8 RTX 4090 GPUs. In the data processing pipeline, we disable mosaic augmentation to better balance the difficulty of the camouflaged object detection task. We adopt an 80-epoch training schedule with a lr of $2 \times 10^{-4}$.

\subsection{Text Pipeline and Fusion}
For sub-description text fusion in the baseline methods, we adopt a simple yet effective strategy of summing the text feature vectors followed by normalization:
\begin{equation}
\label{equation-a1}
\mathbf{t}_c^{\mathrm{fused}} = 
\frac{\sum_{k=1}^{K} \mathbf{t}_{c,k}}{\left\| \sum_{k=1}^{K} \mathbf{t}_{c,k} \right\|_2 }.
\end{equation}
$t_{c,k}$ denotes the embedding vector of the $k$-th sub-description text in category $c$, and $t_c^{\mathrm{fused}}$ represents the fused sub-description text for category $c$. Empirically, this operation preserves sub-description information more effectively and yields improved detection performance. 

% While SDDF employs CLIP for online test-set sub-description encoding, other baselines utilize their native encoders to synthesize category embeddings, while strictly maintaining their original post-processing. 
YOLO-World and DOSOD leverage CLIP to extract sub-description vectors, then the fused category embeddings subsequently passed to a Vision-Language PAN and an MLP adapter, respectively. Similarly, YOLOE utilizes MobileCLIP to generate fused embeddings for its auxiliary network. In contrast, GLIP-T, GLIPv2-T, and Grounding DINO-T employ BERT for text extraction, integrating the fused results into the GLIP fusion encoder or the GroundingDINO neck. To ensure experimental consistency, all models are trained with an identical random seed and a learning rate (lr) of $2 \times 10^{-4}$.

In terms of training configuration, we set the \texttt{lr\_mult} of the MLP-based text adapter in SDDF to 0.01, while assigning an \texttt{lr\_mult} of 0.5 to both the backbone and PAN. We also fix the random seed to the same value across all experiments to ensure fair and reproducible comparisons.

\subsection{Numerical Stability of SF-GLU}
We ensure stable convergence via Epsilon Clamping ($\epsilon=10^{-8}$), Argmax Gradient Detach, and Sigmoid-bounded gain $(1, 1+\alpha)$ with $\alpha=1$.

% ==========================================
% Section C: Additional Quantitative
% ==========================================
\section{Additional Quantitative Results}
\label{sec:add_quant}

\subsection{Comparison with State-of-the-Art (SOTA) COD Methods}
Mainstream COD methods primarily focus on segmentation tasks. To bridge the gap between segmentation-based COD and bbox detection, we train SOTA models on the OVCOD-D under open-set setting. During inference, detection results are derived from the axis-aligned bounding boxes of predicted and ground-truth masks, with localization performance evaluated using the AP shown in Table~\ref{tab:horizontal_comparison}.

% \begin{table}[h]
% \centering
% \caption{Comparison with SOTA COD Methods.}
% \tabcolsep=5pt 
% \resizebox{1\columnwidth}{!}{ 
% % \scriptsize
% % \small
% \begin{tabular}{l|cccc|c} 
% \hline
% Metric & SINet-V2~\cite{sinet} & FSPNet~\cite{fspnet}  & CamoFormer~\cite{camoformer} & HDPNet~\cite{hdpnet}  & \textbf{SDDF-L} \\
% \hline
% $\mathrm{AP}$      & 40.2 & 47.9 & 55.6 & 56.3          & \textbf{56.4} \\
% $\mathrm{AP}_{50}$ & 69.3 & 76.2 & 80.2 & \textbf{81.5} & 76.4 \\
% $\mathrm{AP}_{75}$ & 39.4 & 49.4 & 59.0 & 59.6          & \textbf{60.7} \\
% \hline
% \end{tabular}
% }
% \label{tab:horizontal_comparison}
% \end{table}

\begin{table}[h]
\centering
\caption{Comparison with SOTA COD methods on OVCOD-D.}
\label{tab:horizontal_comparison}
\tabcolsep=12pt % 适当加大列间距，因为列数变少了
\resizebox{1\columnwidth}{!}{ 
\begin{tabular}{l|ccc} 
\hline
Method & $\mathrm{AP}$ & $\mathrm{AP}_{50}$ & $\mathrm{AP}_{75}$ \\
\hline
SINet-V2~\cite{21PAMI-Concealed}      & 40.2 & 69.3 & 39.4 \\
FSPNet~\cite{fspnet}       & 47.9 & 76.2 & 49.4 \\
CamoFormer~\cite{camoformer} & 55.6 & 80.2 & 59.0 \\
HDPNet~\cite{hdpnet}       & 56.3 & \textbf{81.5} & 59.6 \\
\hline
\textbf{SDDF-L (Ours)}    & \textbf{56.4} & 76.4 & \textbf{60.7} \\
\hline
\end{tabular}
}
\end{table}

\subsection{Analysis across Specific Camouflage Difficulty Levels}
To balance the impact of target scale on visual similarity, we define a difficulty metric by weighting the target-image CLIP cosine similarity (0.6) and a scale coefficient $1 - \mathrm{A}_{\mathrm{bbox}}/\mathrm{A}_{\mathrm{img}}$ (0.4), $\mathrm{A}_{\mathrm{bbox}}$ and $\mathrm{A}_{\mathrm{img}}$ denote target and image areas, respectively. Table~\ref{tab:difficulty_analysis} presents results across three quantile-split levels: Mild, Moderate, and Severe.

\begin{table}[h]
\centering
\caption{Multi-level camouflability performance of SDDF-S on OVCOD-D.}
\tabcolsep=4pt
\resizebox{0.8\columnwidth}{!}{
\scriptsize
% \small
\begin{tabular}{l|ccc}
\hline
Method & $\mathrm{AP}_{\mathrm{mild}}$ & $\mathrm{AP}_{\mathrm{mod}}$ & $\mathrm{AP}_{\mathrm{severe}}$ \\
\hline
SDDF-S & 64.2 & 49.8 & 39.0 \\
\hline
\end{tabular}
}
\label{tab:difficulty_analysis}
\end{table}

% ==========================================
% Section D: Extended Ablation Studies
% ==========================================
\section{Extended Ablation Studies}
\label{sec:ablation_ext}

\subsection{Ablation Study on SVD Settings}
SVD is performed on zero-centered matrices $\mathrm{T} \in \mathbb{R}^{K \times D}$. Retained PCs $\in [3, 10]$. 
% Reducing $K$ from 20 to 10 leads to a drop in AP (48.1 $\rightarrow$ 46.1), confirming that semantic richness is key to accurate localization. 
% The 3-layer MLP is trained with lr of $2 \times 10^{-6}$. 
The core function of SVD lies in its ability to eliminate semantic redundancy within sub-descriptions through orthogonal decomposition, thereby rendering different sub-descriptions more semantically independent in the latent space. To quantitatively evaluate this effect, we designed three comparative groups and conducted ablation studies atop the DOSOD baseline, systematically investigating the impact of varying the number of retained maximum principal components in SVD on model detection performance.

As show in Table~\ref{tab:svd_ablation}, employing ten principal components significantly improves model performance, achieving a 2.0 AP increase over the baseline. Reducing the number of components to five leads to substantial semantic information loss, resulting in performance degradation compared to the ten-component configuration. Conversely, increasing the number to fifteen retains excessive redundancy among sub-descriptions, also yielding diminished performance relative to the optimal ten-component setting. Thus, ten principal components strike the ideal balance between preserving semantic integrity and eliminating redundancy.

\begin{table}[h]
\centering
\caption{Ablation study on different maximum number of components obtainable from SVD on the OVCOD-D dataset. All model variants adopt the fusion approach defined in Equation~\ref{equation-a1}. The coverage auxiliary loss and SF-GLU are disabled.}
\label{tab:svd_ablation}
\begin{tabular}{lccc}
\toprule
SVD Configuration & AP & AP\textsubscript{50} & AP\textsubscript{75} \\
\midrule
w/o SVD  & 44.8 & 67.4 & 46.8 \\
w/ SVD (5)  & 46.5 & \textbf{69.8} & 48.3 \\
w/ SVD (10) & \textbf{46.8} & 69.5 & \textbf{49.2} \\
w/ SVD (15) & 46.1 & 69.2 & 48.4 \\
\bottomrule
\end{tabular}
\end{table}

\subsection{Ablation Study on SF-GLU Convolutional Layers}
SF-GLU uses a single-layer $1 \times 1$ convolution to expand text-visual similarity to the same dimension as visual features. Table~\ref{tab:conv_layers_ablation} shows that increasing depth degrades performance, confirming that single-layer adaptation is effective.

\begin{table}[h]
\centering
\caption{Ablation study on the number of convolutional layers.}
\tabcolsep=4pt
\resizebox{0.8\columnwidth}{!}{ 
\scriptsize
\begin{tabular}{c|ccc}
\hline
Conv Layers & $\mathrm{AP}$ & $\mathrm{AP}_{50}$ & $\mathrm{AP}_{75}$ \\
\hline
1 & \textbf{48.1} & \textbf{70.7} & \textbf{50.3} \\
2 & 47.1 & 69.6 & 49.8 \\
4 & 46.0 & 68.7 & 48.6 \\
\hline
\end{tabular}
}
\label{tab:conv_layers_ablation}
\end{table}

% ==========================================
% Section E: Additional Qualitative
% ==========================================
\section{Additional Qualitative Results}
\label{sec:qualitative}

\subsection{Qualitative Examples}
\begin{figure*}[h]
    \centering
    \includegraphics[width=\linewidth]{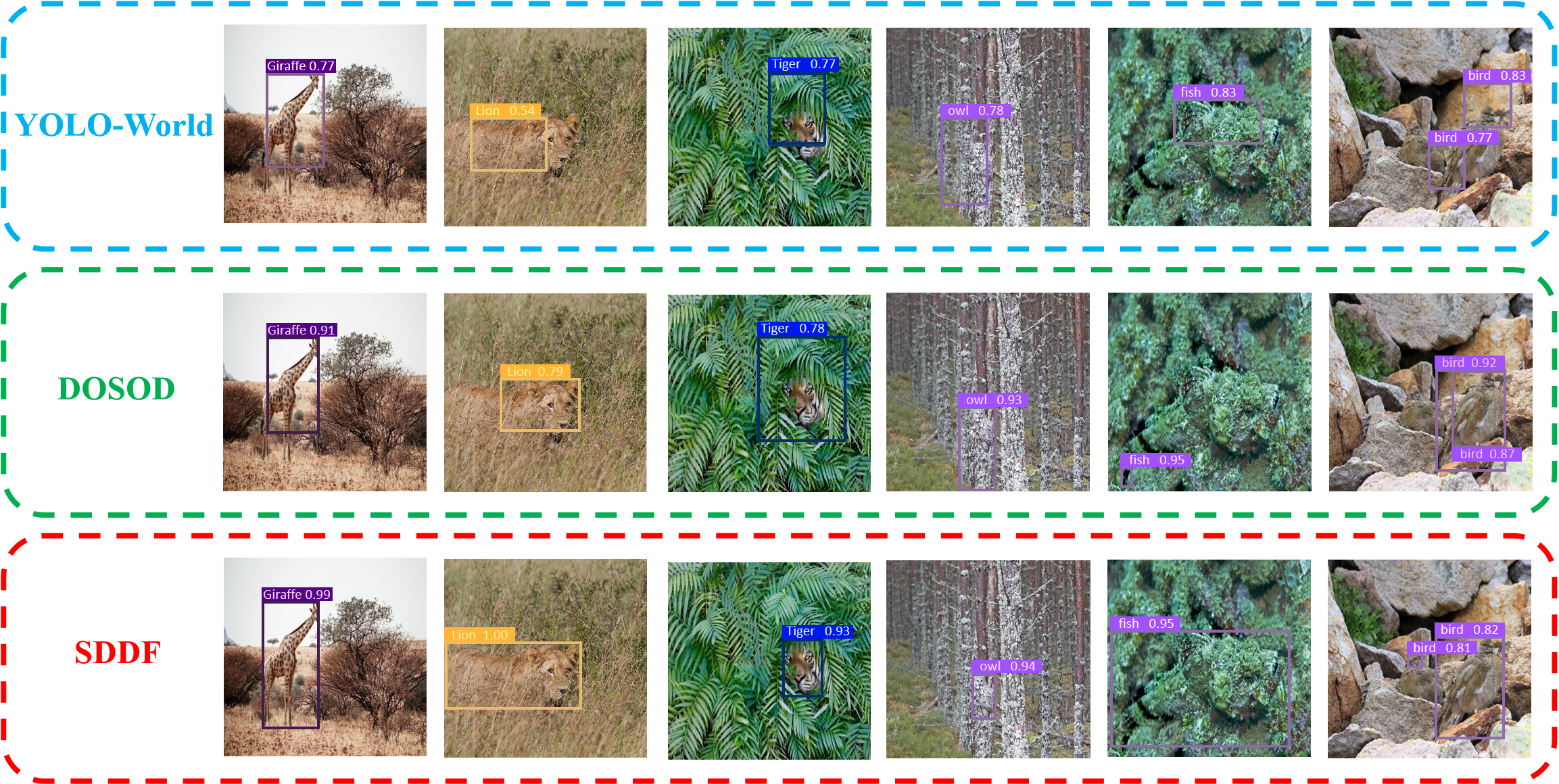}
    \caption{Quantitative comparison is conducted via visualization of the detection bounding boxes.}
    \label{fig:qual_example}
\end{figure*}

As evidenced by the visualizations in the Figure~\ref{fig:qual_example}, the proposed SDDF method significantly outperforms the other two comparative models in precisely localizing camouflaged objects, successfully detecting regions that exhibit high visual similarity with their surrounding environments—regions that often prove challenging for competing approaches.

Furthermore, when dealing with small and medium sized targets (e.g., Owl and Tiger in the fourth and third columns of Figure~\ref{fig:qual_example}, respectively), SDDF exhibits markedly stronger robustness to interference, effectively avoiding the common pitfall of misclassifying background elements as foreground objects. In complex scenes containing multiple targets (e.g., Bird in the sixth column of Figure~\ref{fig:qual_example}), SDDF also demonstrates superior localization accuracy and clearer separation among instances, yielding more reliable and precise bounding-box predictions overall.

\subsection{Attention Improvement of SF-GLU}
\begin{figure*}[h]
    \centering
    \includegraphics[width=\linewidth]{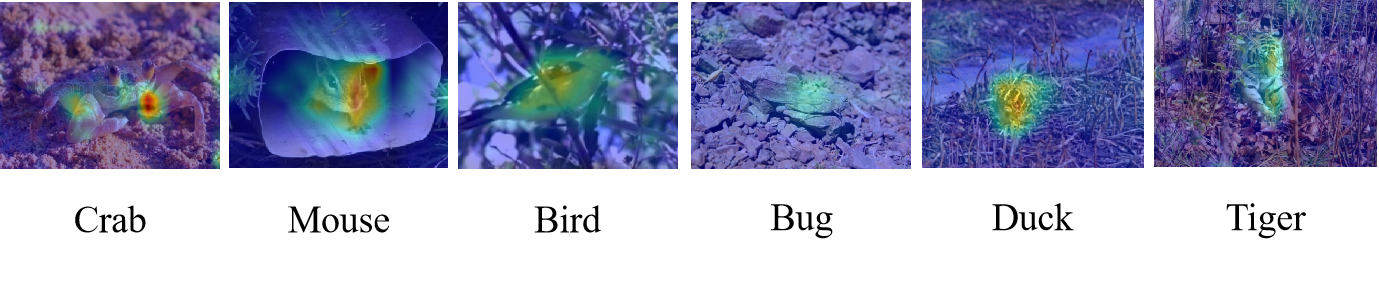}
    \caption{Quantitative comparisons are further conducted by visualizing heatmap representations of the response differences in feature maps with and without the proposed SF-GLU. Specifically, these feature maps are extracted from the PAN outputs.}
    \label{fig:feature_difference_visualze}
\end{figure*}

To investigate the effectiveness of the proposed SF-GLU module in promoting feature focus on camouflaged objects, we visualize heatmap representations of the response differences in feature maps before and after this module.

As demonstrated in the Figure~\ref{fig:feature_difference_visualze}, regions exhibiting high activation (indicated in red and yellow) are precisely concentrated on the camouflaged targets, whereas background regions (marked in blue) show significantly suppressed responses. This highly targeted activation pattern clearly illustrates that, by leveraging specificity-oriented sub-descriptions, the SF-GLU module successfully directs the model’s attention toward regions that exhibit strong semantic alignment with fine-grained descriptors, thereby enabling effective discrimination between camouflaged objects and complex, cluttered backgrounds.

For instance, in the “Mouse” example, the high-response areas are tightly localized to critical parts of the animal, such as the head and body, further underscoring the module’s capability to achieve precise and semantically meaningful feature focusing even under severe camouflage conditions.

%%%%%%%%% REFERENCES
% {
%     \small
%     \bibliographystyle{ieeenat_fullname}
%     \bibliography{main}
% }

\end{document}

%% file: preamble.tex
\ifcsname red\endcsname\else
	
\fi
\ifcsname todo\endcsname\else
	
\fi
\ifcsname TODO\endcsname\else
	
\fi